\def\year{2021}\relax
\documentclass[letterpaper]{article} 
\usepackage{aaai21}  
\usepackage{times}  
\usepackage{helvet} 
\usepackage{courier}  
\usepackage[hyphens]{url}  
\usepackage{graphicx} 
\usepackage{natbib}  
\usepackage{caption} 
\nocopyright

\usepackage{array}
\usepackage{adjustbox}
\usepackage{soul}
\usepackage{url}
\usepackage[utf8]{inputenc}
\usepackage{graphicx}
\usepackage{amsmath}
\usepackage{amsthm}
\usepackage{latexsym,amssymb}
\usepackage{booktabs}
\usepackage{subcaption}
\usepackage{algorithm2e}
\usepackage[dvipsnames]{xcolor}
\usepackage[english]{babel}
\usepackage{xspace}
\usepackage{xfrac}
\usepackage{setspace}
\usepackage{stmaryrd}
\usepackage{enumitem}
\usepackage{verbatim}
\usepackage[all]{xy}
\usepackage{multirow}
\usepackage[switch]{lineno}

\let\subparagraph\paragraph
\usepackage{titlesec}
\let\subparagraph\llncssubparagraph

\definecolor{tyellow1}{HTML}{FCE94F}
\definecolor{tyellow2}{HTML}{EDD400}
\definecolor{tyellow3}{HTML}{C4A000}
\definecolor{torange1}{HTML}{FCAF3E}
\definecolor{torange2}{HTML}{F57900}
\definecolor{torange3}{HTML}{C35C00}
\definecolor{tbrown1}{HTML}{E9B96E}
\definecolor{tbrown2}{HTML}{C17D11}
\definecolor{tbrown3}{HTML}{8F5902}
\definecolor{tgreen1}{HTML}{8AE234}
\definecolor{tgreen2}{HTML}{73D216}
\definecolor{tgreen3}{HTML}{4E9A06}
\definecolor{tblue1}{HTML}{729FCF}
\definecolor{tblue2}{HTML}{3465A4}
\definecolor{tblue3}{HTML}{204A87}
\definecolor{tpurple1}{HTML}{AD7FA8}
\definecolor{tpurple2}{HTML}{75507B}
\definecolor{tpurple3}{HTML}{5C3566}
\definecolor{tred1}{HTML}{EF2929}
\definecolor{tred2}{HTML}{CC0000}
\definecolor{tred3}{HTML}{A40000}
\definecolor{tlgray1}{HTML}{EEEEEC}
\definecolor{tlgray2}{HTML}{D3D7CF}
\definecolor{tlgray3}{HTML}{BABDB6}
\definecolor{tdgray1}{HTML}{888A85}
\definecolor{tdgray2}{HTML}{555753}
\definecolor{tdgray3}{HTML}{2E3436}

\usepackage[pdftex%
,colorlinks=true       
,linkcolor=tblue2        
,citecolor=tblue2        
,filecolor=black       
,urlcolor=tblue2         
,linktoc=page           
,plainpages=false]{hyperref}
\addto\extrasenglish{%

}

\graphicspath{{plots/}}
\DeclareGraphicsExtensions{.pdf}

\titlespacing{\section}{0pt}{*3.15}{*1.125}
\titlespacing{\subsection}{0pt}{*2.25}{*0.9}
\titlespacing{\paragraph}{0pt}{*0.75}{1em}

\makeatletter
\def\thm@space@setup{%
  \thm@preskip=2pt \thm@postskip=2pt
}
\makeatother

\titleformat{\paragraph}[runin]
            {\normalfont\normalsize\bfseries}{\theparagraph}{0.4em}{}

\title{Optimal Decision Lists using SAT}

\author{Jinqiang Yu, Alexey Ignatiev, Pierre {Le Bodic}, Peter J. Stuckey \\}
\affiliations{
  Faculty of Information Technology, Monash University, Australia \\
  \href{mailto:jyuu0044@student.monash.edu}{jyuu0044@student.monash.edu} \\
   \{\mailtodomain{alexey.ignatiev},\mailtodomain{pierre.lebodic},\mailtodomain{peter.stuckey}\}@monash.edu

}
\newcommand{\pjs}[1]{\textcolor{blue}{\textsc{Pjs:} #1}}
\newcommand{\aign}[1]{\textcolor{blue}{\textsc{aign:} #1}}
\newcommand{\plb}[1]{\textcolor{blue}{\textsc{plb:} #1}}

\newcommand{\ignore}[1]{}
\renewcommand{\plb}[1]{}

\newcommand{\fml}[1]{{\ensuremath{\mathcal{#1}}}}

\newcommand{\mailtodomain}[1]{\href{mailto:#1@monash.edu}{\nolinkurl{#1}}}

\makeatletter
\def\thm@space@setup{%
  \thm@preskip=2pt \thm@postskip=2pt
}
\makeatother

\titleformat{\paragraph}[runin]
            {\normalfont\normalsize\bfseries}{\theparagraph}{0.4em}{}

\newcommand{\false}{\textit{false}}

\newcommand{\dr}[2]{\textsf{if~} #1 \textsf{~then~} #2}

\newcommand{\blue}{\color{tblue3}}
\newcommand{\purple}{\color{tred3}}
\newcommand{\green}{\color{tgreen3}}
\newcommand{\grey}{\color{Gray}}

\newcommand*\patchAmsMathEnvironmentForLineno[1]{%
  \expandafter\let\csname old#1\expandafter\endcsname\csname #1\endcsname
  \expandafter\let\csname oldend#1\expandafter\endcsname\csname end#1\endcsname
  \renewenvironment{#1}%
     {\linenomath\csname old#1\endcsname}%
     {\csname oldend#1\endcsname\endlinenomath}}%
\newcommand*\patchBothAmsMathEnvironmentsForLineno[1]{%
  \patchAmsMathEnvironmentForLineno{#1}%
  \patchAmsMathEnvironmentForLineno{#1*}}%
\AtBeginDocument{%
\patchBothAmsMathEnvironmentsForLineno{equation}%
\patchBothAmsMathEnvironmentsForLineno{align}%
\patchBothAmsMathEnvironmentsForLineno{flalign}%
\patchBothAmsMathEnvironmentsForLineno{alignat}%
\patchBothAmsMathEnvironmentsForLineno{gather}%
\patchBothAmsMathEnvironmentsForLineno{multline}%
}

\graphicspath{{plots/}}
\DeclareGraphicsExtensions{.pdf}

\newcounter{example}
\newenvironment{example}
{\refstepcounter{example}\begin{trivlist}\item\textbf{Example~\theexample}}{\qed\end{trivlist}}

\begin{document}

\maketitle{}

\begin{abstract}
Decision lists are one 
of the most easily explainable machine learning models. 
Given the renewed emphasis on explainable machine learning decisions, 
this machine learning model is increasingly attractive, combining small size and clear explainability.
In this paper, we show for the first time how to construct optimal ``perfect'' decision lists which are perfectly accurate on the training data, and minimal in size, making use of modern SAT solving technology.
We also give a new method for determining optimal sparse decision lists, which trade off size and accuracy.
We contrast the size and test accuracy of optimal decisions lists versus optimal decision sets, as well as other state-of-the-art methods for determining optimal decision lists. 
We also examine the size of average explanations generated by 
decision sets and decision lists.
\end{abstract}

\section{Introduction} \label{sec:intro}

With the increasing use of Machine Learning (ML) models to automate decisions, 
there has been an upsurge in interest in 
\emph{explainable artificial intelligence} (XAI) 
where these models can explain, in a manner understandable by humans, why they made a decision.
This in turn has led to a re-examination of machine learning models that are implicitly easy to explain.

Arguably the most explainable forms of ML models are
decision trees, decision lists and decision sets,
since these encode simple logical rules.
Of these, decision sets provide the simplest explanation, since if a rule ``fires'' for a given data instance, then this rule is the only explanation required.
To explain decision lists, i.e.\ ordered sets of decision rules, we additionally need to take the order of rules into account. 
\ignore{
Decision lists are ordered sets of decision rules.
To explain decisions made by a decision list, we need to give not only the rule that fires, but also all the earlier rules that did not fire.
To explain the decisions of a decision tree, we need to show the path in the tree taken by the instance to the decision. 
}

In order to make explanations easy for humans to understand they should be as small as possible.
There has been considerable investigation into producing the
smallest possible optimal perfect decision trees~\cite{nipms-ijcai18,verhaeghe2019learning}, where the decision tree agrees perfectly with the training data, 
as well as producing the smallest possible sparse decision trees~\cite{hu2019optimal,dl85}, where there is a trade-off between size of the decision tree versus its accuracy on the training data.
Similarly there has been investigation of smallest possible optimal decision sets~\cite{ipnms-ijcar18},
as well as recent work on optimal sparse decision sets~\cite{yisb-corr20}. 
This body of work provides compelling evidence that optimal sparse machine learning models generalize very well, providing high testing accuracy.

Recent work on optimizing decision lists~\cite{rudin-kdd17a,rudin-mpc18,JMLR:v18:17-716} relies on a two-phase approach. First, decision rules are mined using some association rule mining technique~\cite{itemset-mining}, then an optimal order of the rules is found via search. 
In contrast, the method proposed in this paper directly generates all the rules of the optimal decision list as part of the search.
This means it can generate decision rules which are meaningful in the context of their position in the target decision list, but could by themselves not provide valuable information on the training data, and therefore would not have been mined by the approach of~\cite{rudin-kdd17a,rudin-mpc18,JMLR:v18:17-716}.
The two step process can also generate many candidate rules to order when the number of features is large.
Finally, this earlier approach does not optimize rules in terms of reducing the number of literals, which may result in larger decision list models.

The previous methods focus on sparse decision lists, that trade size of the decision list for accuracy on the training data.
In contrast, we can also find optimal perfect decision lists, which agree completely with the training data.

In summary the contributions of this paper are:
\begin{itemize}
    \item The first method to determine optimal perfect decision lists, that agree with all the training data.
\ignore{    
    \item A more scalable approach to sparse optimal decision lists when the number of features is large (but the number of examples is small)\pjs{No evidence!}
}    
    \item An approach to optimal sparse decision lists that generates more accurate decision lists than previous methods.
    \item We introduce the notion of \emph{explanation size} which, for a particular example, determines how much information is required to explain the classification given by a machine learning model. We compare explanation size for decision lists and decision sets.
\end{itemize}

\ignore{
The rest of paper is organized as follows.
In the next section we introduce preliminaries about MaxSAT and decision rules and lists.
In Section~\ref{sec:relw} we examine related work.
In Section~\ref{sec:enc} we show how to encode optimal decision lists for perfect and sparse problems in MaxSAT. 
In Section~\ref{sec:sep} we examine how we can make the approaches more scalable by decomposing the problem into one problem per class. 
In Section~\ref{sec:size} we discuss how can we judge how succinct an explanation is to an example, using a decision set or decision list.  This gives us a new way of judging how explainable a decision list or set is.
In Section~\ref{sec:res} we give experimental results.
Finally in Section~\ref{sec:conc} we conclude.
}

\section{Preliminaries} \label{sec:prelim}

\paragraph{(Maximum) Satisfiability.}
The input of a Boolean satisfiability problem (SAT)~\cite{sat-handbook09} consists of a formula over a set of propositional variables using various logic operators on these variables.
Solving a SAT consists in determining whether there exists an assignment of True or False value to each variable, called a \emph{truth assignment}, such that the entire formula is \emph{satisfied}, i.e.\ True.
Otherwise, the formula is \emph{unsatisfiable}.
%
In the more specific \emph{conjunctive normal form} (CNF) form, the formula is a conjunction of clauses, and each clause is a disjunction of \emph{literals}.
A literal is a variable or its negation.
Hence, a CNF formula can be satisfied if and only if at least one literal per clause can be set to True.

In the context of unsatisfiable formulas, the maximum satisfiability (MaxSAT) problem consists in finding a truth assignment that maximizes the number of satisfied clauses.
In the Partial Weighted MaxSAT variant~\cite[Chapter~19]{sat-handbook09}, each clause $c$ is either \emph{soft} and has a weight $w_c$, or it is \emph{hard}.
An optimal solution then consists in a truth assignment that satisfies all hard clauses and maximizes the sum of the weights of the satisfied soft clauses.




%
\paragraph{Classification Problems}
We consider a classification problem with a set of features $\fml{F} = \{f_1,\ldots,f_K\}$ and a label \fml{C}, all binary (or binarized using standard techniques~\cite{scikitlearn-full}).
The training set is denoted $\fml{E}=\{e_1,\ldots,e_M\}$.
Each instance $e_i \in \fml{E}$ is a pair $(c_i, \pi_i) \in \fml{C} \times 2^\fml{F}$.
Given \fml{E}, the classification problem consists in finding a function $\hat{\phi}:2^\fml{F} \rightarrow \fml{C}$ which minimizes the classification error on testing data.



%

\paragraph{Rules, Decision Sets and Decision Lists.}
We can naturally represent a binary feature $f \in \fml{F}$ as a Boolean variable, and its two possible values as a literal or its negation, denoted $f$ and $\neg{f}$.
A \emph{rule} has the form
IF ``instance satisfies formula'' THEN ``predict $c$'',
where the formula is a conjunction on a subset of the feature literals.

A \emph{Decision Set} is an \emph{unordered} set of rules.
A decision set misclassifies an instance if no rule matches the instance, or if at least two rules predict different classes. 

A \emph{Decision List} is an \emph{ordered} set of rules.
The first rule of a decision list that matches an instance is the one that classifies the instance.
Decision list are often written as a single cascade of IF-THEN-ELSEs, with the last rule often a ``catch-all'', or \emph{default rule}, which matches all (remaining) instances to some class.

\begin{example}\label{ex:data}
Consider the following set of 8 items (shown as columns):
\setlength{\tabcolsep}{3pt}
\begin{center}
\begin{tabular}{crrrrrrrrrr}
\toprule
\multicolumn{2}{c}{Item No.} & 1 & 2 & 3 & 4 & 5 & 6 & 7 & 8  \\
\midrule
\multirow{5}{0mm}{\rotatebox{90}{Features}}
& $A$      & 1 & 1 & 0 & 0 & 0 & 0 & 0 & 0  \\
& $B$      & 0 & 1 & 1 & 1 & 0 & 0 & 0 & 0 \\
& $C$      & 1 & 0 & 0 & 1 & 1 & 1 & 0 & 0 \\
& $D$      & 0 & 1 & 1 & 0 & 0 & 1 & 1 & 1 \\
& $E$      & 0 & 1 & 0 & 1 & 1 & 0 & 0 & 1 \\ \midrule 
\multicolumn{2}{r}{Class $H$}
           & 1 & 1 & 0 & 0 & 1 & 1 & 0 & 0 \\
\bottomrule
\end{tabular}
\end{center}

\ignore{ 
\begin{tabular}{crrrrrrrrrr}
\toprule
\multicolumn{2}{c}{Item No.} & 1 & 2 & 3 & 4 & 5 & 6 & 7 & 8  \\
\midrule
\multirow{5}{0mm}{\rotatebox{90}{Features}}
& $A$      & \blue 1 & \blue 1 & 0 & 0 & 0 & 0 & 0 & 0  \\
& $B$      & 0 & 1 & \purple 1 & \purple 1 & 0 & 0 & 0 & 0 \\
& $C$      & 1 & 0 & 0 & 1 & \green 1 & \green 1 & 0 & 0 \\
& $D$      & 0 & 1 & 1 & 0 & 0 & 1 & \grey 1 & \grey 1 \\
& $E$      & 0 & 1 & 0 & 1 & 1 & 0 & 0 & 1 \\ \midrule 
\multicolumn{2}{r}{Class $H$}
           & \blue 1 & \blue 1 & \purple 0 & \purple 0 & \green 1 & \green 1 & \grey 0 & \grey 0 \\
\bottomrule
\end{tabular}
}

A valid and optimal decision set for this data is
\begin{equation*}
\begin{array}{rrcr}
\textsf{if} & A & \textsf{then} & H \\
\textsf{if} & \neg B \wedge C & \textsf{then} & H \\
\textsf{if} & \neg A \wedge \neg C & \textsf{then} & \neg H \\
\textsf{if} & \neg A \wedge B & \textsf{then} & \neg H \\
\end{array}
\end{equation*}
\ignore{
\begin{array}{rrcl}
\blue \textsf{if} & \blue L & \blue \textsf{then} & \blue \neg H \\
\green \textsf{if} & \green \neg L \wedge \neg C & \green \textsf{then} & \green H \\
\purple \textsf{if} & \purple C & \purple \textsf{then} & \purple \neg H \\
\end{array}
}
\ignore{
\begin{eqnarray*}
\blue L & \blue \Rightarrow & \blue \neg H \\
\green \neg L \wedge \neg C & \green \Rightarrow & \green H \\
\purple C & \purple \Rightarrow & \purple \neg H
\end{eqnarray*}
}
The \emph{size} of this decision set is 11 (one for each literal on the left hand and right hand side, or alternatively, one for each literal on the left hand side and one for each rule). Note how rules can overlap: both the first and second rule classify item~1.

A valid and optimal decision list for the data above is
\begin{equation*}
\begin{array}{rl}
& \dr{A}{H} \\
\textsf{else} & \dr{B}{\neg H} \\
\textsf{else} & \dr{C}{H} \\
\textsf{else} & \dr{true}{\neg H} \\
\end{array}
\end{equation*}
\ignore{
\begin{array}{rl}
& \blue \dr{A}{H} \\
\textsf{else} & \purple \dr{B}{\neg H} \\
\textsf{else} & \green \dr{C}{H} \\
\textsf{else} & \grey \dr{true}{\neg H} \\
\end{array}
}
The size of the decision list is 7, and there is no overlap of rules by definition: item 1 is classified by the first rule and not the second. Note how the last rule is a default rule.
\end{example}

\section{Related Work} \label{sec:relw}

Decision lists were introduced by \citeauthor{rivest-ml87}~(\citeyear{rivest-ml87}) and heuristic methods for decision lists also date back to the late 80s~\cite{clark-ml89,clark-ewsl91}. 

One recent approach~\cite{rudin-mpc18} provides DLs that have some optimality guarantee. Given a fixed set of decision rules, it chooses a minimum-size ordered subset of these rules; the order essentially terminates when a default rule is chosen. The authors model the problems as an integer program (IP) and solve it with a mixed integer programming (MIP) solver.  The objective is a combination of training accuracy and sparsity, minimizing misclassifications where every rule used incurs a ``cost'' of $C$ misclassifications, and every literal used costs $C_1$ misclassifications. The method is slow, and somewhat restricted by the time required to generate all potential possible rules as input. They consider data sets with up to 3000 examples and 60 features, but cannot prove optimality of  their solutions on the data tested. One advantage of the approach is that it is easy to customize, for example, favoring the use of certain features, or extending to cost-sensitive learning.

To the best of our knowledge, the first method to generate \emph{optimal decision lists} extends the approach of~\cite{rudin-mpc18} 
using the same idea of ordering a fixed set of decision rules, but using a bespoke branch-and-bound algorithm~\cite{JMLR:v18:17-716}.
The method makes use of bounding methods and 
symmetry elimination techniques. They minimize regularized misclassification, where each rule costs $\lambda M$ misclassification errors
where $M$ is the number of training examples.
The approach relies on the sparsification parameter $\lambda$ to limit the set of rules it needs to consider.
It can find and prove 
optimal solutions to large problems (hundreds of thousands of examples),
the main limitation is on the number of features, since the number of possible decision rules grows exponentially in the number of features. 
The data sets they consider have up to 28 (binary) features, and at most 189 decision rules are considered.


\section{Encoding} \label{sec:enc}

Recent work of \cite{yisb-corr20} gives a SAT encoding for describing decision sets. We can modify this fairly naturally to instead define decision lists. 

\subsection{MaxSAT Model for Perfect Decision Lists} \label{sec:sat}

\newcommand{\noas}[1]{}
\newcommand{\newnoas}[1]{#1}

The~\cite{yisb-corr20}
MaxSAT model determines whether
there exists a perfect \emph{decision set} of at most a 
given size $N$.
All rules are encoded as a single sequence of feature literals, and a class literal ends each rule.
The model also keeps track of which items are valid (i.e.\ agree) with previous literals in the rule.
We can modify this to determine a perfect \emph{decision list} of size at most $N$ by keeping track of which items have previously been classified by a previous rule, and preventing them from being considered (in)valid in later rules. 
Note that for binary classification problems we consider that there is one class pseudo-feature $\fml{C} = \{c\}$ and items have this feature or not.
For 3 or more classes, we assume a one-hot encoding~\cite{scikitlearn-full}, with each item having exactly one feature from $\fml{C}$.

The sequence of literals is viewed as a path graph, with one feature literal per node. 
The encoding uses a number of Boolean variables described below:
\begin{itemize}
   \item $s_{jr}$:$\hspace{3.3pt} $ node $j$ is a literal on feature
      $f_r\in\fml{F}\cup\fml{C}$;
   \item $t_{j}$:$\,\,$\,\,\, truth value of the literal for node $j$;
   \item $v_{ij}$:$\hspace{4.5pt} $ example $e_i\in\fml{E}$ is valid
      at node $j$;
   \item $n_{ij}$:\hspace{2.5pt} example $e_i\in\fml{E}$ is not previously classified by any nodes before $j$
   \item $u_j$:\hspace{6pt} node $j$ is unused
\end{itemize}


The model is as follows:
\begin{itemize}
\item A node either decides a feature or is unused:
\begin{equation}\label{eq:used} \forall_{j \in [N]}\ u_j +
\sum_{r\in[K+|\fml{C}|]} s_{jr} = 1 \end{equation}
\item If a node $j$ is unused then so are all the following nodes:
\begin{equation}\label{eq:newlast} \forall_{j \in [N-1]}\ u_{j}
\rightarrow u_{j+1} \end{equation}
\item The last used node is a leaf:
\begin{eqnarray}
\forall_{j \in [N-1]}\ u_{j+1} \rightarrow u_j \vee \bigvee_{c \in \fml{C}} s_{jc} \\
u_{N} \vee \bigvee_{c \in \fml{C}} s_{Nc}
\end{eqnarray}
   \item All examples are not previously classified at the first node:
      \begin{equation}\label{eq:nfirst}
      \forall_{i \in [M]}\ n_{i1} \end{equation}
   \item An example $e_i$ is previously unclassified at node $j+1$ 
      iff it was previously unclassified, and either $j$ is not a leaf node 
      or it was invalid at the previous leaf node (so not classified by the rule that finished there):
      %
      %
      \begin{equation}\label{eq:nassign}
      \forall_{i \in [M]} \forall_{j \in [N-1]}\ n_{ij+1}
      \leftrightarrow n_{ij} \wedge ((\bigwedge_{c \in \fml{C}} \neg s_{jc}) \vee \neg v_{ij})
      \end{equation}

   \item All examples are valid at the first node:
      \begin{equation}\label{eq:vfirst}
      \forall_{i \in [M]}\ v_{i1} \end{equation}
   \item An example $e_i$ is valid at node $j+1$ iff $j$ is a leaf
      node and it was previously unclassified, 
      or $e_i$ is valid at node $j$ and $e_i$ and node $j$ agree
      on the value of the feature $s_{jr}$ selected for that node:
      %
      %
      \begin{equation}\label{eq:valid}
      \forall_{i \in [M]} \forall_{j \in [N-1]}\ 
       \begin{array}{l} v_{ij+1}
      \leftrightarrow 
      (((\bigvee_{c \in \fml{C}} s_{jc}) \wedge n_{ij+1}) ~\vee \\[1mm]
      (v_{ij} \wedge \bigvee_{r\in[K]}
      {(s_{jr} \wedge (t_j = {\pi_i[r]}))})
     )
      \end{array}
      \end{equation}
   \item If example $e_i$ is valid at a leaf node $j$, it should
      agree on the class feature:
      \begin{equation}\label{eq:class}
      \begin{array}{r@{~~~~}l}
      \forall_{i \in [M]} \forall_{j \in [N]}\ (s_{jc} \wedge v_{ij})
      \rightarrow (t_j = c_i) & \fml{C} = \{c\} \\[3mm]
       \forall_{i \in [M]} \forall_{j \in [N]} \forall_{c \in \fml{C}}\ (s_{jc} \wedge v_{ij})
      \rightarrow c_i & |\fml{C}| \geq 3
      \end{array}
      \end{equation}
   \item When there are 3 or more classes we restrict leaf nodes to only consider \emph{true} examples of the class:
      \begin{equation}\label{eq:true-multi}
      \forall_{j \in [N]}\forall_{c \in \fml{C}}\ s_{jc} \rightarrow t_j \hspace*{1cm} |\fml{C}| \geq 3 
      \end{equation}

   \item For every example there should be at least one leaf node
      where it is valid:
      \begin{equation}\label{eq:correct}
      \forall_{i \in [M]}\ \bigvee_{j\in[N]}((\bigvee_{c \in \fml{C}} s_{jc}) \wedge v_{ij})
      \end{equation}
\end{itemize}
The constraints~\eqref{eq:used}--\eqref{eq:correct} make up the hard constraints of the MaxSAT model.
As for the optimization criterion, we maximize $\sum_{j \in [N]} u_j$,
which can be trivially represented as a list of unit soft clauses of
the form $(u_j, 1)$.

The differences between the above model and the model of~\cite{yisb-corr20} is the addition of the $n_{ij}$ variables to track which items have been previously classified, and their use in constraint~\eqref{eq:valid},
as well as the rules to compute them given in constraints~\eqref{eq:nfirst} and \eqref{eq:nassign}.

The model shown above represents a non-clausal Boolean formula, which
can be clausified with the use of auxiliary
variables~\cite{tseitin68}.
Also note that any of the known cardinality encodings that can be used
to represent the sum in~\eqref{eq:used}~\cite[Chapter~2]{sat-handbook09}
(also see~\cite{asin-sat09,bailleux-cp03,batcher-afips68,sinz-cp05}).
Finally, the size (in terms of the number of literals) of the proposed
SAT encoding is $\fml{O}(N \times M \times K)$, which results from
constraints~\eqref{eq:nassign} and~\eqref{eq:valid}.

\begin{example}\label{ex:soln}
Consider a solution for 7 nodes for the data of \autoref{ex:data}.
The representation of the decision list is shown below:
\begin{equation*}
   \newcommand{\xyo}[1]{*++[o][F-]{#1}}
    \newcommand{\xyoo}[1]{*++[o][F=]{#1}}
 \xymatrix@=4mm{
     1  & 2 & 3 & 4 & 5& 6 & 7 \\
    \xyo{A} \ar[r] & \xyo{H} \ar@{..>}[r] & \xyo{B} \ar[r] & \xyo{\neg H} \ar@{..>}[r]
     & \xyo{C}\ar[r] & \xyo{H}\ar@{..>}[r] & \xyo{\neg H}}
\end{equation*}
The interesting (true) decisions for each node are 
as follows:

$$
\begin{array}{llccccccc}
   \toprule
      \mbox{Node } j & \;\;\;\;\; & 1 & 2 & 3 & 4 & 5 & 6 & 7 \\ \midrule 
s_{jr} & & s_{1A} & s_{2H} & s_{3B} & s_{4H} & s_{5V} & s_{6H} & s_{7H} \\ \midrule
t_{j}  & & 1  & 1  & 1 & 0 & 1  & 1  & 0 \\ \midrule
v_{ij} & & v_{11}  & v_{12} & v_{33} & v_{34} & v_{55} & v_{56} & v_{77} \\
       & & \vdots & v_{22} & \vdots & v_{44}  & \vdots  & v_{66} & v_{87} \\
       & &  v_{81} &   & v_{83} & & v_{85} &   &   \\
       \midrule
n_{ij} && n_{11} & n_{12} & n_{33} & n_{34} & n_{55} & n_{56} & n_{77} \\
       && \vdots & \vdots & \vdots & \vdots & \vdots  & \vdots  & n_{87}    \\
       && n_{81} & n_{82} & n_{83} & n_{84} & n_{85}  & n_{86} & \\

   \bottomrule
\end{array}
$$
Note how at the end of each rule, the selected variable is the class $H$.
Note that at the start and after each leaf node all previously unclassified examples are valid, and
each feature literal reduces the valid set for the next node.
In each leaf node $j$ the valid examples are of the correct class determined by
the truth value $t_j$ of that node.
\end{example}

The MaxSAT model tries to find a decision set of size at most $N$.
If this fails, we can increase $N$ by some amount and resolve, until either resource limits (typically computation time) are reached or a solution is found.

\ignore{
\subsection{MaxSAT Model for Mixed Decision Sets and Lists} \label{sec:sat}

We now design a SAT model which determines whether
there exists a mixed decision set or list of at most a given size $N$.

The encoding only differs from that for decisions lists of Section~\ref{sec:sat} by the introduction of a decision $x_j$ which specifies if $j$ is a leaf node whether it ends a decision list, and therefore the next node is start of a new decision list. In that sense if $x_j$ is \false{} then there is an else following the current rule.

The encoding uses the extra Boolean variables described below:
\begin{itemize}
   \item $x_j$: $\,\,$\,\,\,node $j$ ends a decision list
\end{itemize}

The model adds the following equations:

\begin{itemize}
   \item The last used node ends a decision list:
      \begin{equation}\label{eq:lastx}
      \forall_{j \in [N-1]} u_{j+1} \rightarrow u_j \vee x_j
      \end{equation}
      \begin{equation}\label{eq:lastx}
      u_{N} \vee x_N
      \end{equation}

    \item An end node is always a leaf:
      \begin{equation}\label{eq:leafx}
      x_j \rightarrow \bigvee_{c \in \fml{C}} s_{jc}
      \end{equation}
\end{itemize}

The only other change is to replace Equation~\eqref{eq:nassign} by
a version extended to use the $x_j$ variable. 

\begin{itemize}

   \item An example $e_i$ is previously unclassified at node $j+1$ 
      iff $j$ is an end of decision list node, or it was previously unclassified, and either $j$ is a not leaf node 
      or it was invalid at the previous leaf node (so not classified by the rule that finished there):
      %
      %
      \begin{equation}\label{eq:xnassign}
      \forall_{i \in [M]} \forall_{j \in [N-1]}\ n_{ij+1}
      \leftrightarrow x_j \vee (n_{ij} \wedge ((\bigwedge_{c \in \fml{C}} \neg s_{jc}) \vee \neg v_{ij}))
      \end{equation}

\end{itemize}

\begin{example}
Consider the data set of Example~\ref{ex:data}. 
A valid mixed decision list for the class  $H$ is
$$
\begin{array}{rrcl}
\blue \textsf{if} & \blue L & \blue \textsf{then} & \blue \neg H \\
\purple \textsf{if} & \purple C & \purple \textsf{then} & \purple \neg H \\
\green \textsf{else if} & \green \neg L          &  \green \textsf{then}              & \ \green H
\end{array}
$$
There are two decision lists, the first one is a single rule, the second one consists of two rules.
The size of the mixed decision list is 6, and now there overlap of rules by definition: items 4 and 6 are both classified by the first rule and the second. 

The representation of the rules, as a sequence of nodes is shown below:
$$
   \newcommand{\xyo}[1]{*++[o][F-]{#1}}
    \newcommand{\xyoo}[1]{*++[o][F=]{#1}}
 \xymatrix@=4mm{
     1  & 2 & 3 & 4 & 5& 6 & 7 \\
    \xyo{L} \ar[r] & \xyo{\neg H} & \xyo{C} \ar[r] & \xyo{\neg
     H} \ar@{..>}[r]  & \xyo{\neg L} \ar[r]& \xyo{H} & \xyo{~}}
$$
The interesting (true) decisions for each node are given in the following
table

$$
\begin{array}{llccccccc}
   \toprule
      & \;\;\;\;\; & 1 & 2 & 3 & 4 & 5 & 6 & 7 \\ \midrule 
s_{jr} & & s_{1L} & s_{2H} & s_{3C} & s_{4H} & s_{5L} & s_{6H} & u_7 \\ \midrule
t_{j}  & & 1  & 0  & 1 & 0 & 0  & 1  & - \\ \midrule
x_{j}  & & 0  & 1  & 0 & 0 & 0  & 1  & - \\ \midrule

v_{ij} & & v_{11}  & v_{12} & v_{13} & v_{44} & v_{15} & v_{36} & \\
       & & \vdots & v_{22} & \vdots & v_{64} & v_{25}  & v_{56} &  \\
       & &  v_{81} &  v_{42}  & v_{83} & v_{74} & v_{35} & v_{86}  &   \\
       & &  &  v_{62} &   & & v_{55}  &  & \\
       & &  &         &   & & v_{85} \\
       \midrule
n_{ij} && n_{11} & n_{12} & n_{13} & n_{14} & n_{15} & n_{16} & \\
       && \vdots & \vdots & \vdots & \vdots & n_{25} & n_{26}  &    \\
       && n_{81} & n_{82} & n_{83} & n_{84} & n_{35} & n_{36} \\
       &&        &        &  & & n_{55} & n_{56} \\
       &&        &        &  & & n_{85} & n_{86} \\
   \bottomrule
\end{array}
$$
\end{example}

}
\subsection{MaxSAT Model for Sparse Decision Lists} \label{sec:sparse}

We can extend the MaxSAT model
to look for sparse decisions lists that are accurate for most of the instances, rather than perfect.
We minimize the number of misclassifications (including non-classifications, where no decision rule
in the list gives information about the item) plus the size of the decision list
in terms of nodes multiplied by a discount factor $\Lambda$ which records that
$\Lambda$ fewer misclassifications are worth the addition of one node to the decision list.
Typically we define $\Lambda = \lceil \lambda M \rceil$, where $\lambda$ is the regularized cost
of nodes in terms of misclassifications.

We introduce variable $m_i$ to represent that example $i \in [M]$ is misclassified.
The model is as follows:
\begin{itemize}
\item If example $e_i$ is valid at a leaf node $j$ then they agree on the
  class feature or the item is misclassified:
      \begin{equation}\label{eq:misclassified}
      \begin{array}{r@{~~~~}l}
      \forall_{i \in [M]} \forall_{j \in [N]}\ (s_{jc} \wedge v_{ij})
      \rightarrow (t_j = c_i \vee m_i) & \fml{C} = \{c\} \\[3mm]
       \forall_{i \in [M]} \forall_{j \in [N]} \forall_{c \in \fml{C}}\ (s_{jc} \wedge v_{ij})
      \rightarrow (c_i \vee m_i) & |\fml{C}| \geq 3
      \end{array}
      \end{equation}
\item For every example there should be at least one leaf literal
  where it is valid or the item is misclassified (actually \emph{non-classified}):
  \begin{equation}\label{eq:unclassified}
  \forall_{i \in [M]}\
  m_i \vee \bigvee_{j\in[N]}( \bigvee_{c \in \fml{C}} s_{jc} \wedge v_{ij})
  \end{equation}
\end{itemize}
together with all the hard constraints of the model for perfect decision lists except constraints~\eqref{eq:class} and~\eqref{eq:correct}.
The objective function is
\begin{equation*}
\sum_{i \in [M]} m_i + \sum_{j \in [N]} \Lambda (1 - u_j) + N \Lambda
\end{equation*}
represented as soft clauses $(\neg m_i, 1)$, $i\in[M]$,
and $(u_j, \Lambda)$, $j\in[N]$.

\section{Separated Models} \label{sec:sep}

A convenient feature of minimal decision sets is the following: the union of 
minimal decision sets for each $c \in \fml{C}$
that correctly classifies all instances of class $c$ and do not misclassify any instances not
of class $c$ as class $c$, 
is a minimal decision set for the entire problem.

That means we can compute perfect decision sets for $\vert \fml{C} \vert$ classes by separately computing $\vert \fml{C} \vert$ perfect decision sets, one for each class.
The union of these $\vert \fml{C} \vert$ models, which we call \emph{``separated model''}, is clearly not much smaller than the complete model, as
a separated model still covers each example.
The advantage is that computing $\vert \fml{C} \vert$ models of total size $N$ is much faster than computing a single model of size $N$.

For decision lists this property no longer holds.
If we compile decision lists separately for each class,
we must still order the decision lists of different classes.
And it may be that no optimal decision list 
can be expressed as rules for one class, followed by another class, followed by another.

\begin{example}\label{ex:separated}
Consider the dataset of \autoref{ex:data}. Recall that an optimal decision list shown in \autoref{ex:data} has 7 literals.
\ignore{
\setlength{\tabcolsep}{3pt}
\begin{center}
\begin{tabular}{crrrrrrrrrr}
\toprule
\multicolumn{2}{c}{Item No.} & 1 & 2 & 3 & 4 & 5 & 6 & 7 & 8  \\
\midrule
\multirow{5}{0mm}{\rotatebox{90}{Features}}
& $A$      & \blue 1 & \blue 1 & 0 & 0 & 0 & 0 & 0 & 0  \\
& $B$      & 0 & 1 & \purple 1 & \purple 1 & 0 & 0 & 0 & 0 \\
& $C$      & 1 & 0 & 0 & 1 & \green 1 & \green 1 & 0 & 0 \\
& $D$      & 0 & 1 & 1 & 0 & 0 & 1 & \grey 1 & \grey 1 \\
& $E$      & 0 & 1 & 0 & 1 & 1 & 0 & 0 & 1 \\ \midrule 
\multicolumn{2}{r}{Class $H$}
           & \blue 1 & \blue 1 & \purple 0 & \purple 0 & \green 1 & \green 1 & \grey 0 & \grey 0 \\
\bottomrule
\end{tabular}
\end{center}

An optimal decision list for the data above is

$$
\begin{array}{rl}
& \blue \dr{A}{H} \\
\textsf{else} & \purple \dr{B}{\neg H} \\
\textsf{else} & \green \dr{C}{H} \\
\textsf{else} & \grey \dr{true}{\neg H} \\
\end{array}
$$
}
An optimal decision list that is separated in class order is 
\begin{equation*}
\begin{array}{rl}
& \dr{A}{H} \\
\textsf{else} & \dr{\neg B \wedge C}{H} \\
\textsf{else} & \dr{B}{\neg H} \\
\textsf{else} & \dr{true}{\neg H} \\
\end{array}
\end{equation*}
\ignore{
\begin{array}{rl}
& \blue \dr{A}{H} \\
\textsf{else} & \green \dr{\neg B \wedge C}{H} \\
\textsf{else} & \purple \dr{B}{\neg H} \\
\textsf{else} & \grey \dr{true}{\neg H} \\
\end{array}
}
requiring one more literal.
\end{example}

Given that separated models are important for scaling this approach to larger problems, we need to consider approaches for defining decision lists in a separated form.
We consider a number of different approaches:
\begin{description}
    \item[fixed $\sigma$] Given a permutation $\sigma$ of classes, 
    find an optimal decision list for the first class in $\sigma$, then make an optimal decision list for the second class ignoring items already classified by the decision list for the first class. Then consider the third class, etc.
\item[greedy] Make an optimal decision list for each class independently: choose the one that is best under some metric. Fix its solution as the first part of the decision list. Calculate $I'$ as the items not classified by this decision list. Make an optimal decision list for each remaining class independently. Again, choose the best one and fix it. Continue until all classes are considered, or $I'$ becomes empty. 
\end{description}
For the fixed permutation case, one can try all possible permutations, if there are not too many, e.g. $|\fml{C}| \leq 3$, or use a heuristic to choose a permutation $\sigma$. 
One heuristic we consider is sorting the classes by increasing/decreasing number of their respective items in the training set.
Alternatively, we consider ordering the classes greedily based on the post-hoc analysis of the accuracy or cost of individual class representations obtained on the training data.
Here, training accuracy for the representation of class $c$ is $ 1 - \frac{\sum_{i \in [M], c_i = c} m_i}{|\{e_i \in \fml{E}, c_i=c\}|}$ while the cost of representation of class $c$ is assumed to be $N - \sum_{j \in [N]} u_j + \left\lceil\frac{\sum_{i \in [M]} m_i}{ \Lambda}\right\rceil$.

Note that for separated sparse models, the objective is effectively different. Using the same objective for each class separately means that we count a misclassification once for every class it is detected by. This is arguably more informative. As we cannot guarantee the same optimal solutions anyway (due to order restrictions), this seems acceptable.

\section{Explanation Size}\label{sec:size}

Given two different ML models, we can ask \emph{which model gives the smallest explanation} on a particular data instance.  
By optimizing the size of a decision list or decision set, we believe the size of the explanations it creates will be small, but this is not completely accurate.  The explanation size of an ML model can be far smaller than the whole model.
The implicit notion of \emph{explanation size} we are trying to capture is, if a customer/user were to ask why our model made a decision for their case, how would we explain that decision?
Note that we also define explanation size for the cases where a decision set makes no decision, either since no rule fires, or two contradictory rules fire.
We define the explanation size of a model $\hat{\phi}$ on an example instance $e$ as follows.

If $\hat{\phi}$ is a decision set and the rules in $\hat{\phi}$ that fire on example $e$ are $\{\textbf{if~} \pi_i \textbf{~then~} c_i\}$, $\forall i \in [n]$, $n\leq N=|\hat{\phi}|$, then
\begin{itemize}
    \item if all the classes $c_1, \ldots, c_n$ agree, i.e.\ $c_i=c'$, $\forall i\in [n]$, $c' \in \fml{C}$, then the explanation size for example $e$ is $\frac{\sum_{i=1}^n |\textbf{if~} \pi_i \textbf{~then~} c_i|}{n}$, that is, the average of the rules, any of which could explain the example.
    \item if not all classes agree for $e$ then the explanation size is the sum of averages of the rules for all the conflicting classes predicted for $e$; wlog.\ assume that $c_i=c'$, $c' \in \fml{C}$, $1 \leq i \leq k < n$ and $c_j=c''$, $c'' \in \fml{C}$, $k+1\leq j \leq n$, then the explanation size for example $e$ is $\frac{\sum_{i=1}^k |\textbf{if~} \pi_i \textbf{~then~} c_i|}{k} + \frac{\sum_{j=k+1}^n |\textbf{if~} \pi_j \textbf{~then~} c_j|}{n-k}$; similar reasoning can be applied to situations of more than two conflicting classes.
    \item if no rule fires then the explanation size is $|\hat{\phi}|$, i.e. we need the whole decision set to explain why $e$ is not classified.
\end{itemize}

If $\hat{\phi}$ is a decision list and $\textbf{if~} \pi_j \textbf{~then~} c_j$ is the first rule
in $\hat{\phi}$ that fires on example $e$ then
\begin{itemize}
    \item the 
    explanation size is $\sum_{i=1}^j |\pi_i| + 1$ as we need to explain why none of the previous rules fired, and why rule $j$ did.
    \item if no rule fires for $e$ then the explanation size is
    $|\hat{\phi}|$, i.e. we need the whole model to explain why $e$ is not classified. Note that in practice this does not occur since the last rule will be a default rule, and all examples will be classified.
\end{itemize}

Note that it is easy to extend the notion of explanation size to decision trees (as the path from root to leaf) though decision tree models are not considered in this paper.

\ignore{
We can extend this notion to decision trees straightforwardly, if the example $e$ reaches a leaf node $n$ then the explanation size is 
$k$ where $k$ is number of nodes traversed, since each internal node is effectively a literal, and the final node defines the class.
\pjs{Do we want this since the we probably dont compare against them}
\aign{We don't compare but I think we can mention this as a general observation.}}

\section{Experimental Results} \label{sec:res}

\begin{figure*}[!t]
  \centering
  \begin{subfigure}[b]{0.3\textwidth}
    \centering
    \includegraphics[width=\textwidth]{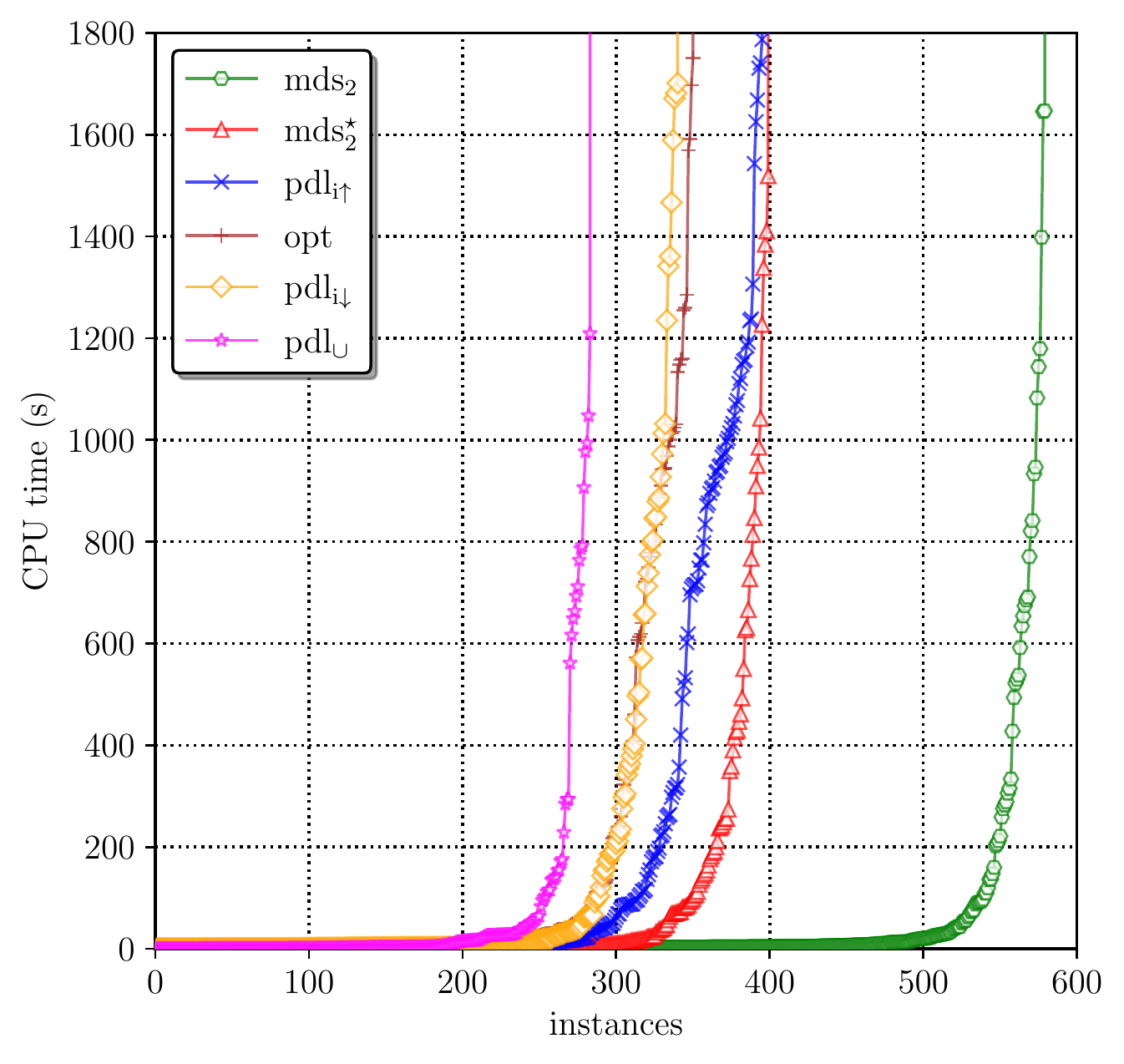}
    \caption{Performance}
    \label{fig:prtime}
  \end{subfigure}%
  \begin{subfigure}[b]{0.3\textwidth}
    \centering
    \includegraphics[width=\textwidth]{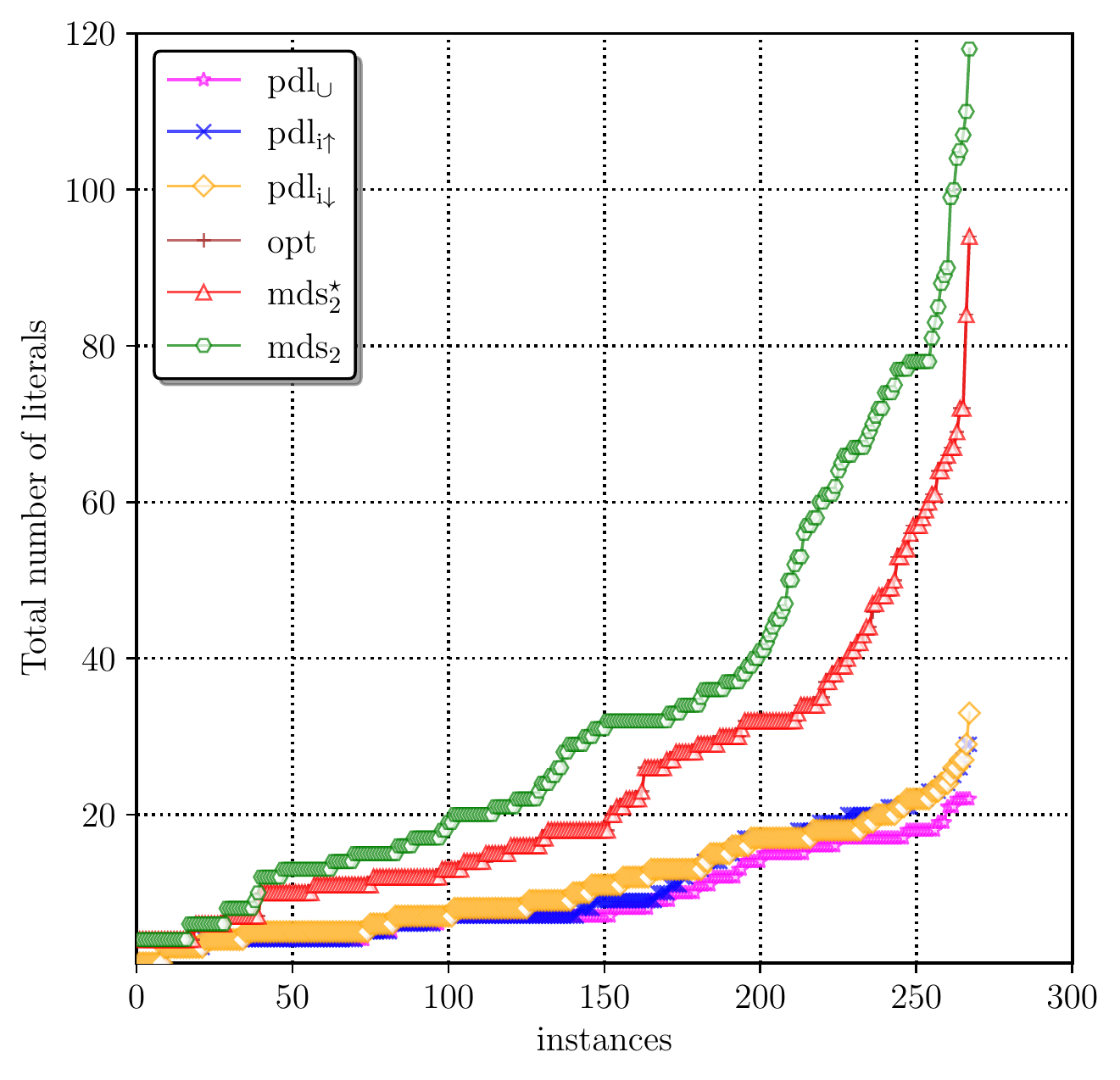}
    \caption{Model size}
    \label{fig:pmsize}
  \end{subfigure}%
  \begin{subfigure}[b]{0.3\textwidth}
    \centering
    \includegraphics[width=\textwidth]{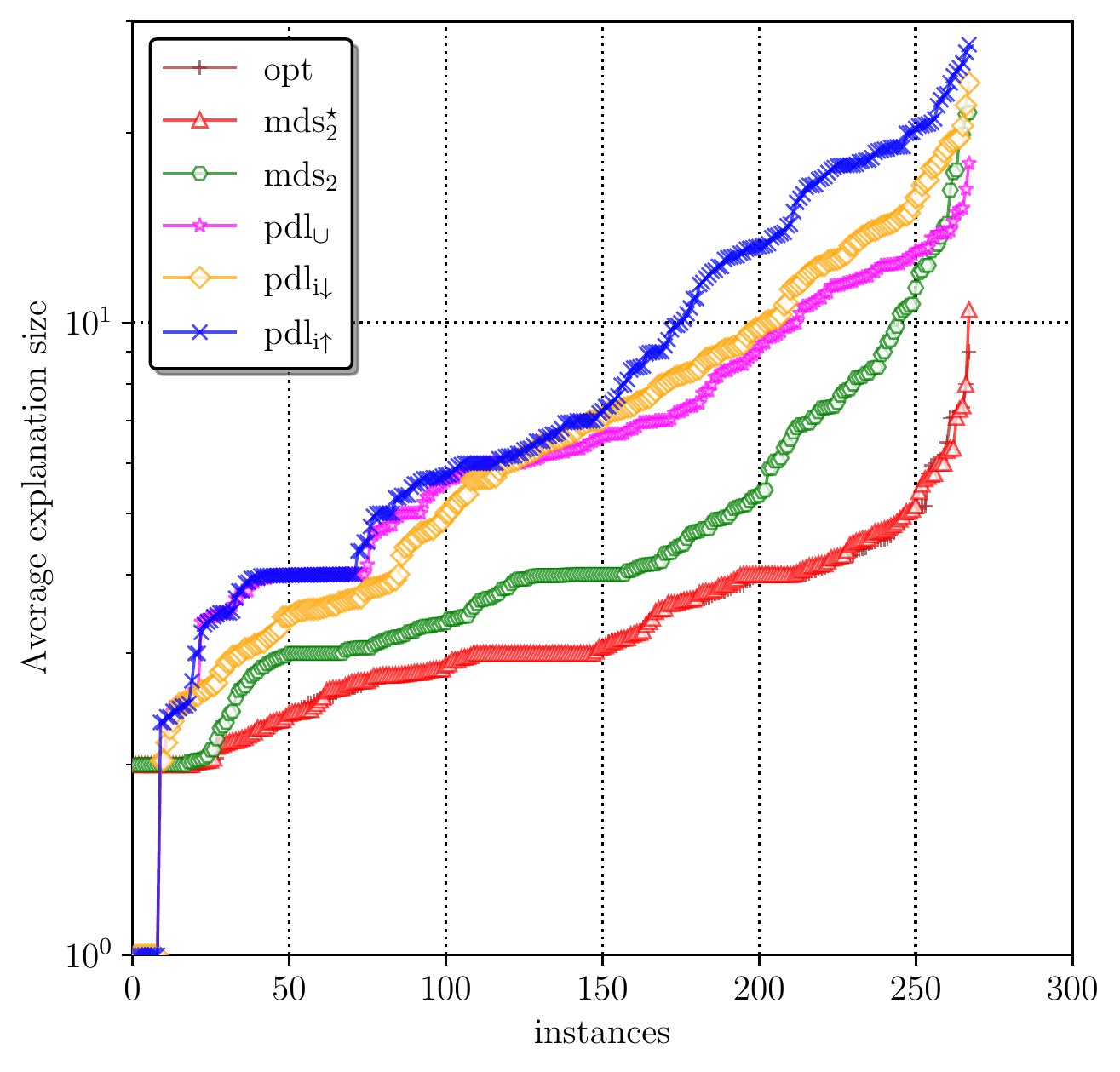}
    \caption{Average explanation size}
    \label{fig:pxsize}
  \end{subfigure}
  
  \begin{subfigure}[b]{0.3\textwidth}
    \centering
    \includegraphics[width=\textwidth]{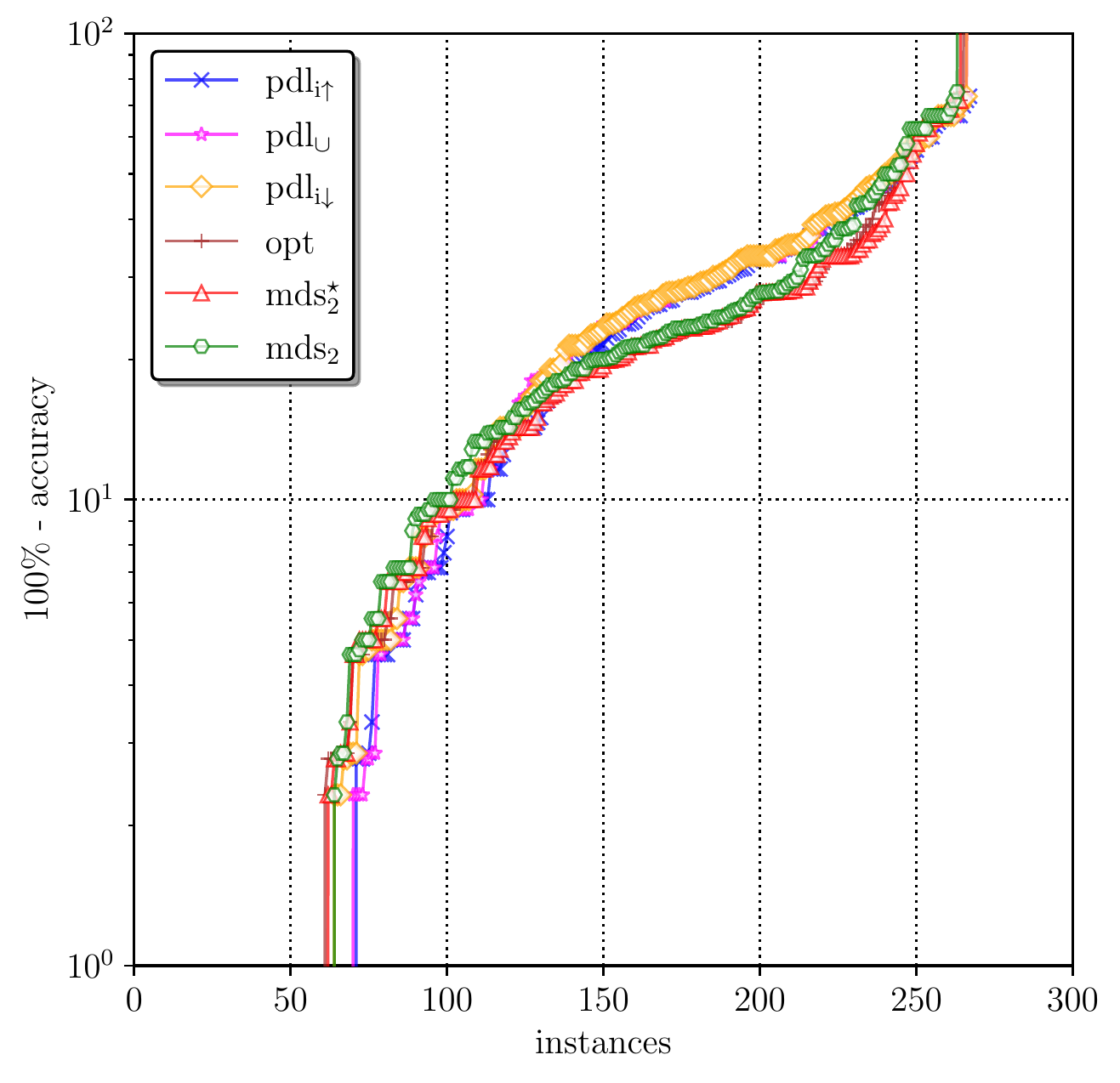}
    \caption{Test accuracy}
    \label{fig:paccy}
  \end{subfigure}
  \begin{subfigure}[b]{0.3\textwidth}
    \centering
    \includegraphics[width=\textwidth]{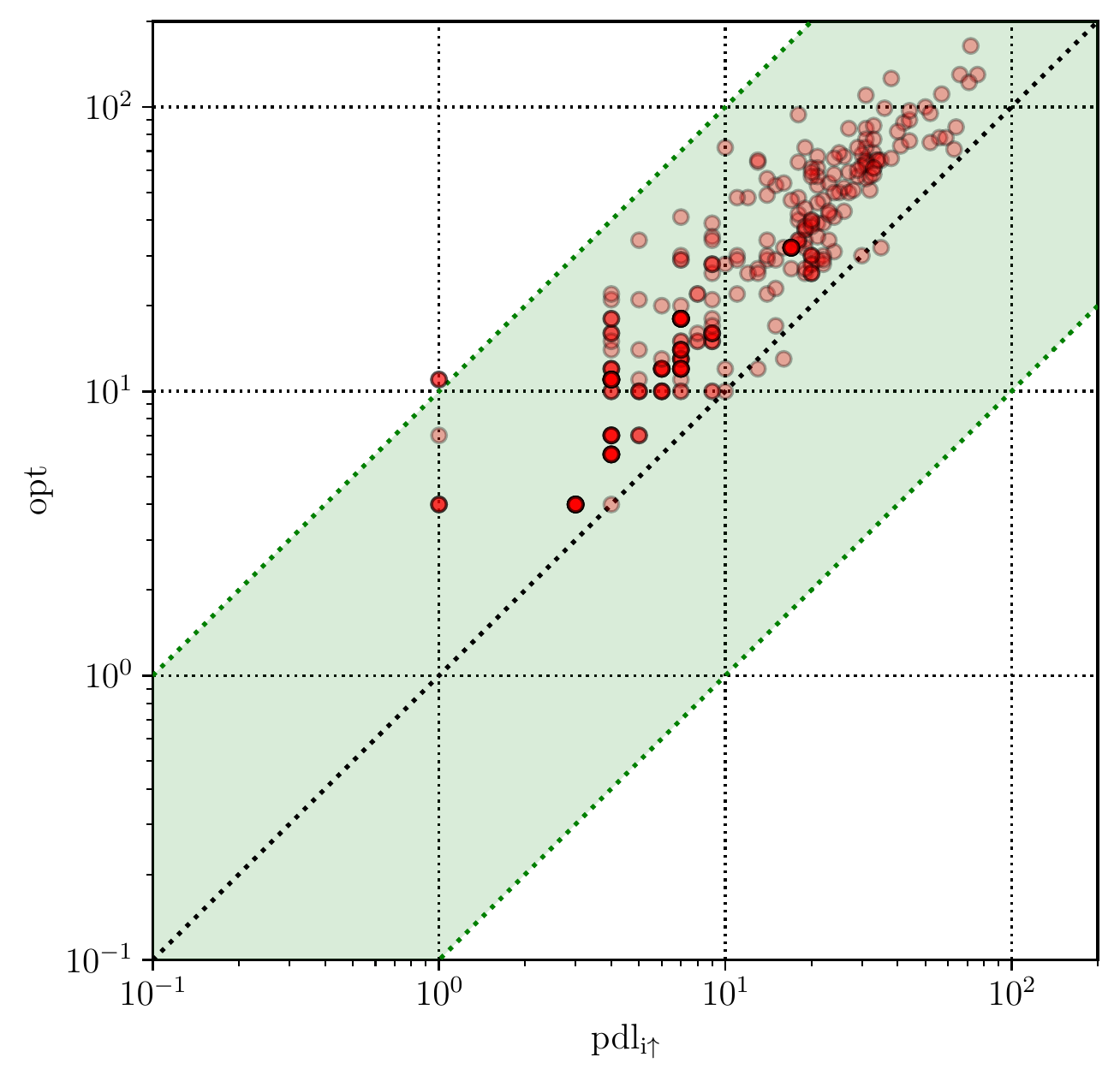}
    \caption{Model size detailed: pdl$_{i\uparrow}$ vs.~opt}
    \label{fig:pmsize-det}
  \end{subfigure}%
  \begin{subfigure}[b]{0.3\textwidth}
    \centering
    \includegraphics[width=\textwidth]{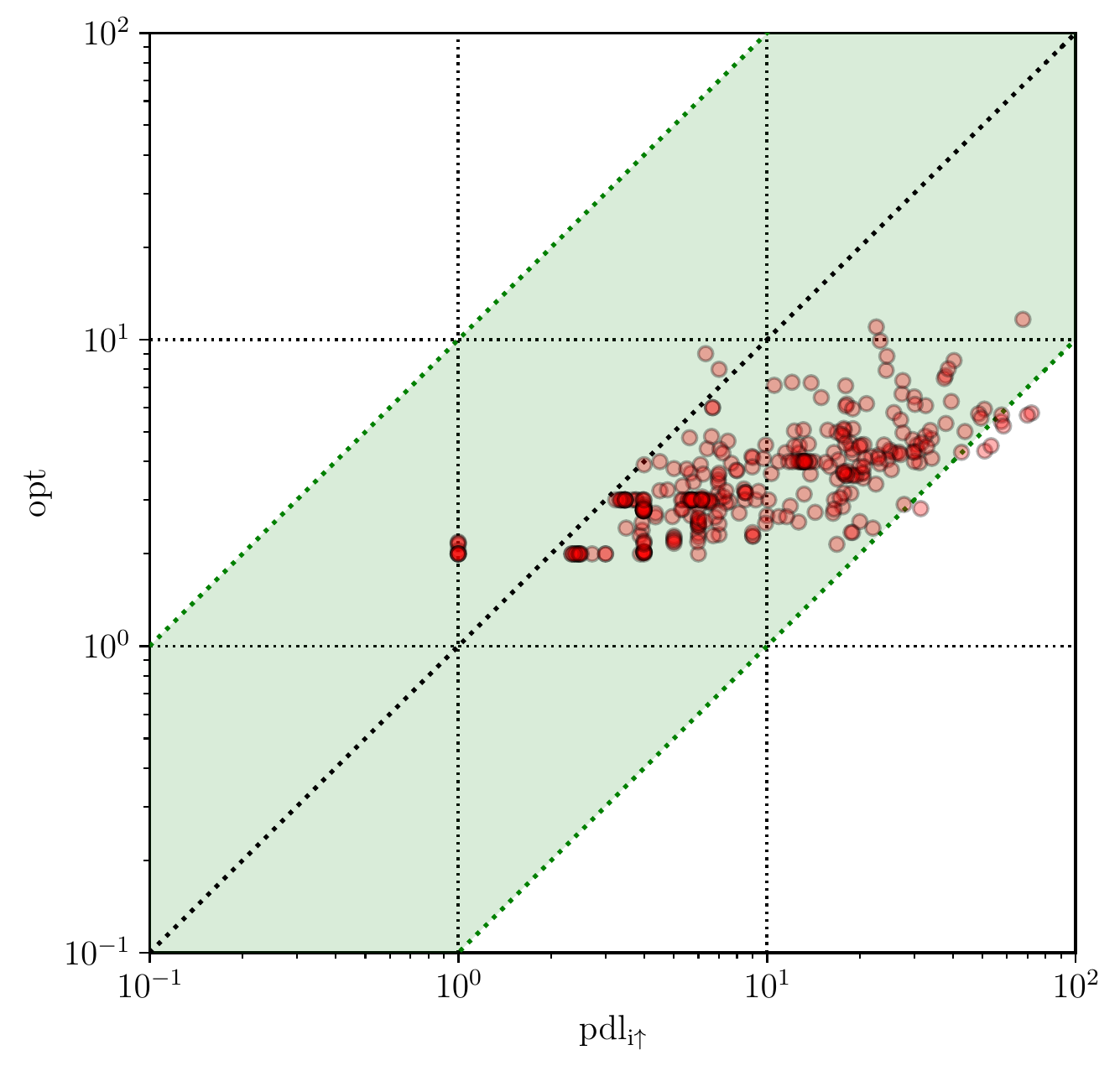}
    \caption{Explanation size detailed: pdl$_{i\uparrow}$ vs.~opt}
    \label{fig:pxsize-det}
  \end{subfigure}
  \caption{Comparison of perfect models.}
  \label{fig:perfect}
\end{figure*}

\begin{figure*}[!t]
  \centering
  \begin{subfigure}[b]{0.3\textwidth}
    \centering
    \includegraphics[width=\textwidth]{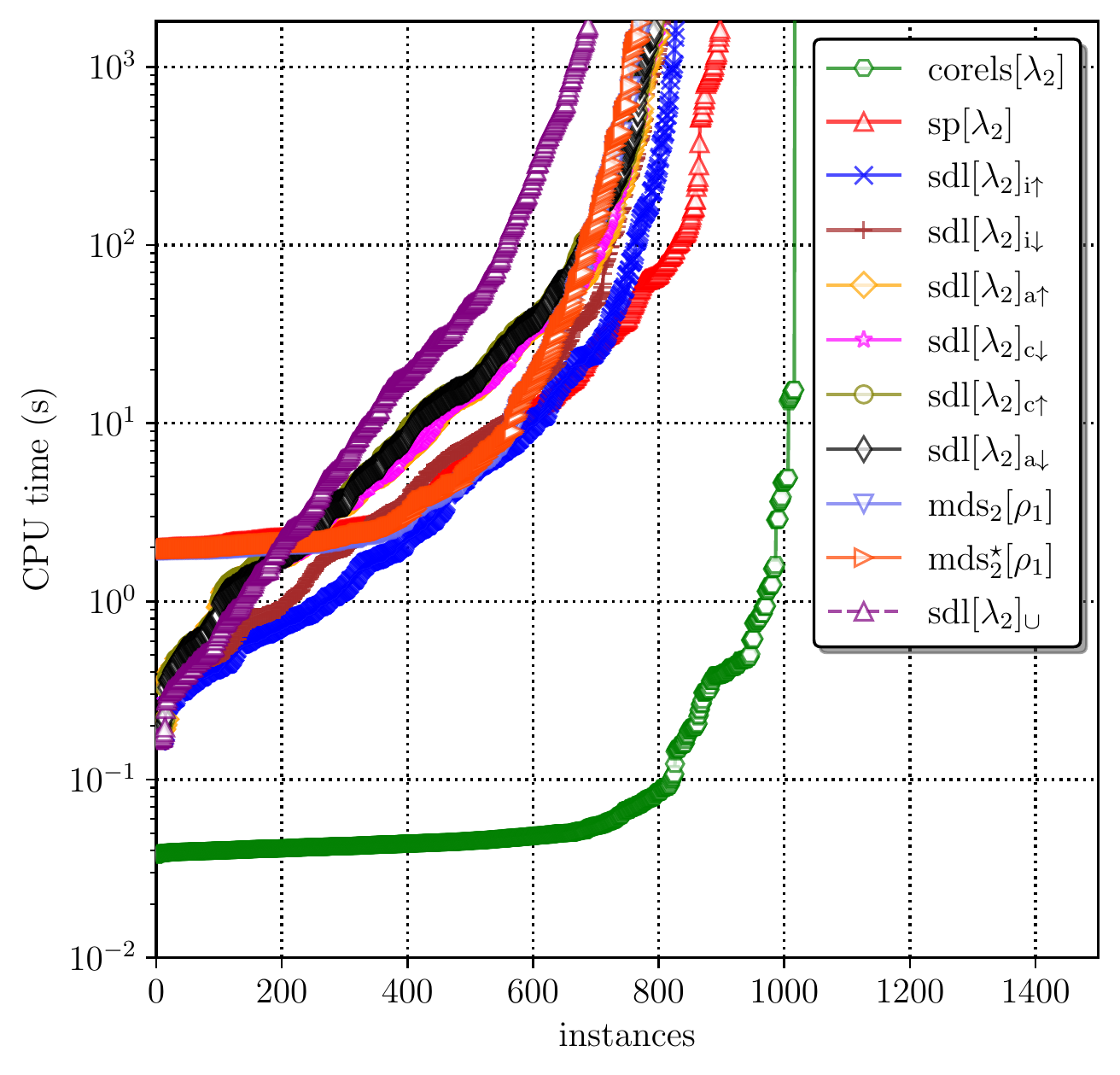}
    \caption{Performance}
    \label{fig:srtime}
  \end{subfigure}%
  \begin{subfigure}[b]{0.3\textwidth}
    \centering
    \includegraphics[width=\textwidth]{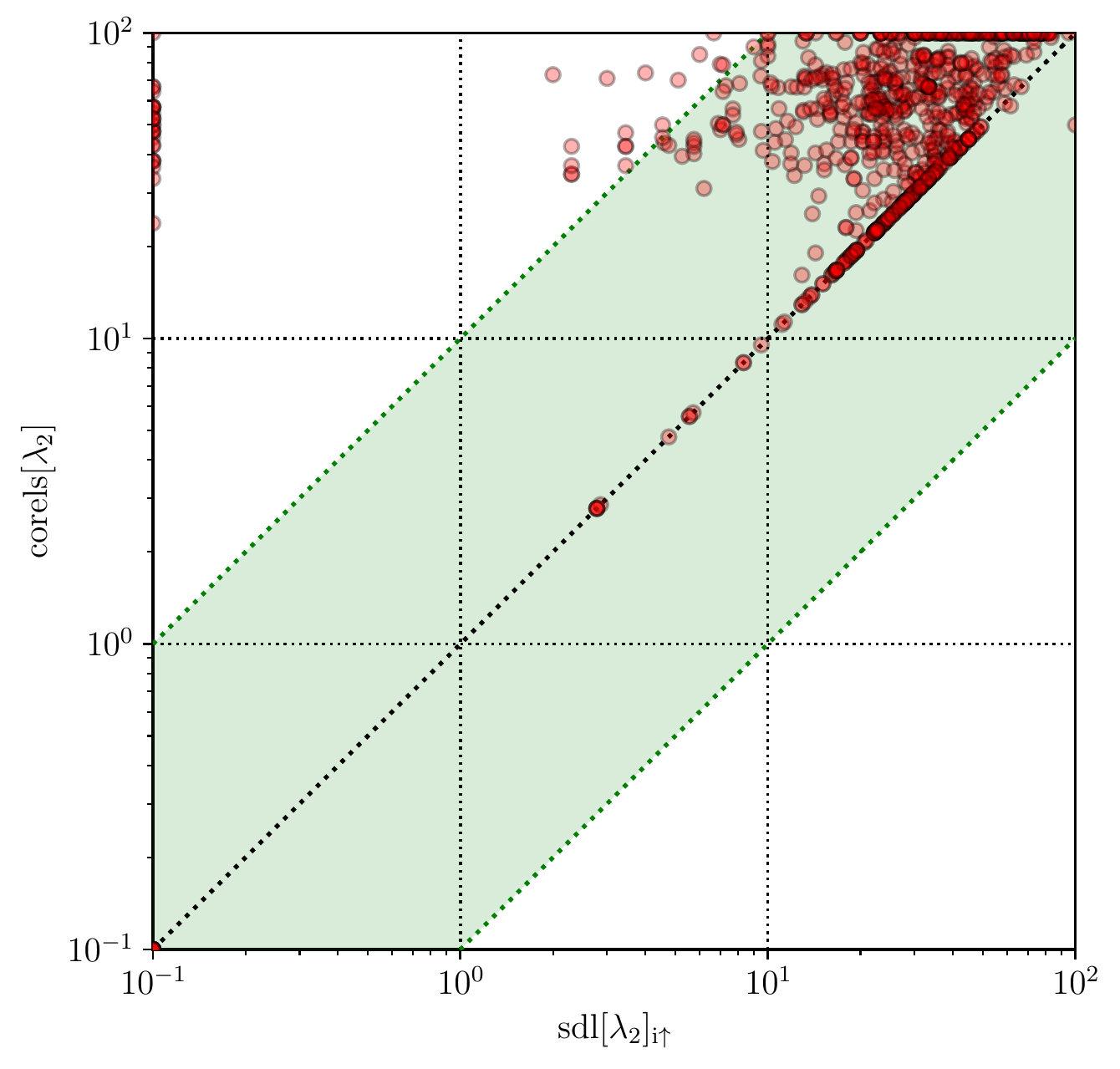}
    \caption{Error: sdl$[\lambda_2]_{i\uparrow}$ vs.~corels$[\lambda_2]$}
    \label{fig:saccy-corels}
  \end{subfigure}%
  \begin{subfigure}[b]{0.3\textwidth}
    \centering
    \includegraphics[width=\textwidth]{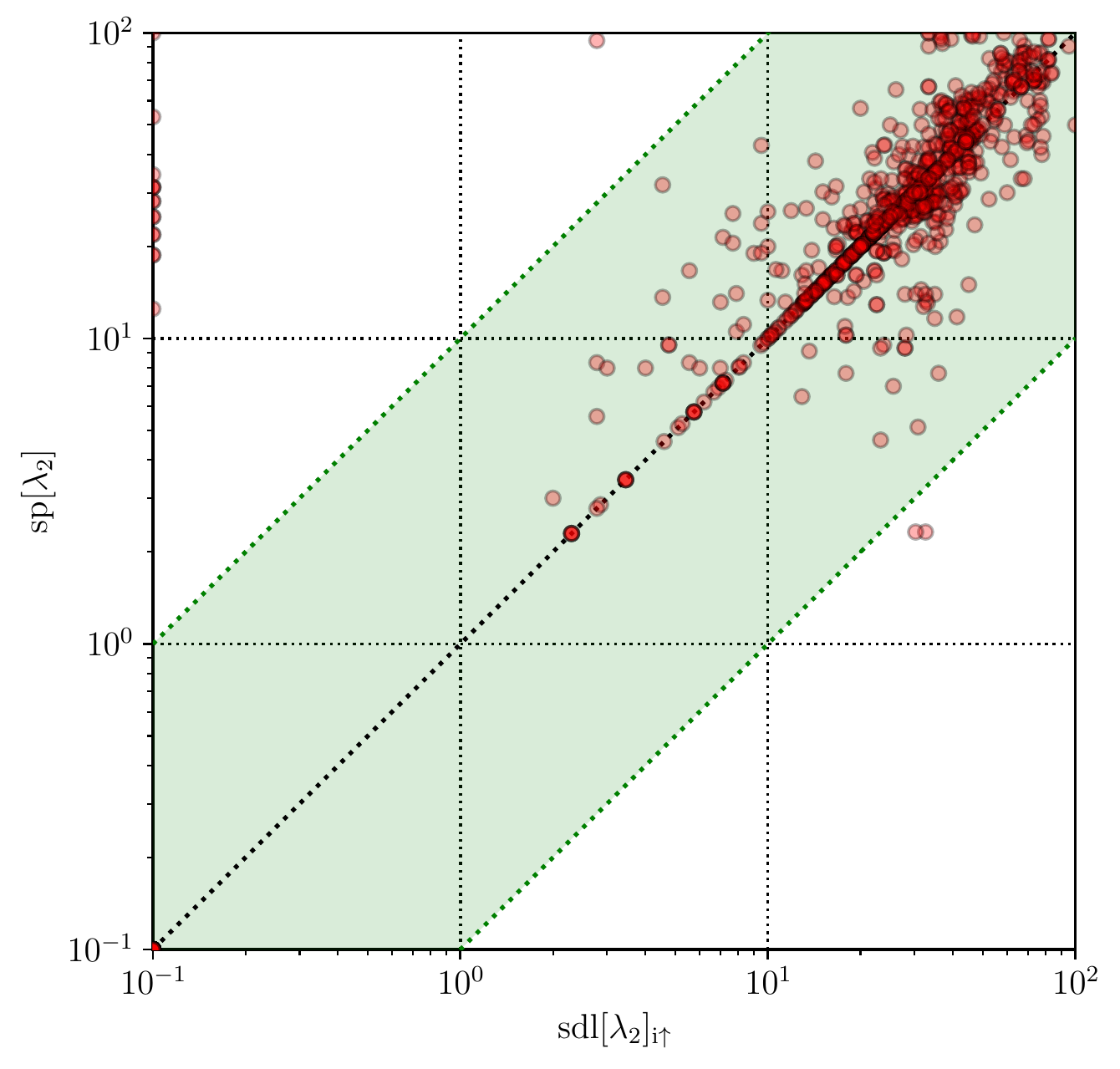}
    \caption{Error: sdl$[\lambda_2]_{i\uparrow}$ vs.~sp$[\lambda_2]$}
    \label{fig:saccy-dset}
  \end{subfigure}
  
  \begin{subfigure}[b]{0.3\textwidth}
    \centering
    \includegraphics[width=\textwidth]{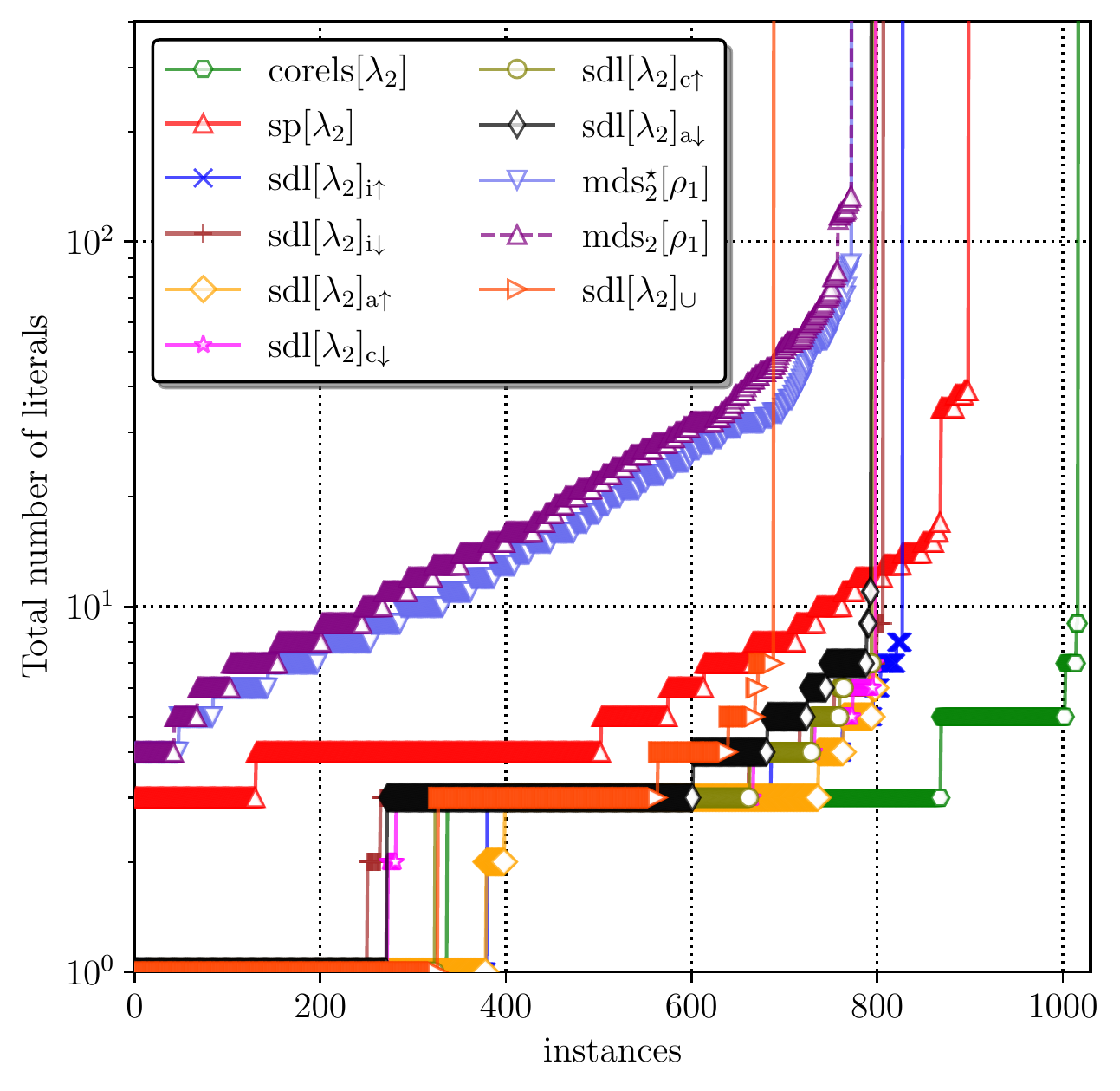}
    \caption{Model size}
    \label{fig:smsize}
  \end{subfigure}
  \begin{subfigure}[b]{0.3\textwidth}
    \centering
    \includegraphics[width=\textwidth]{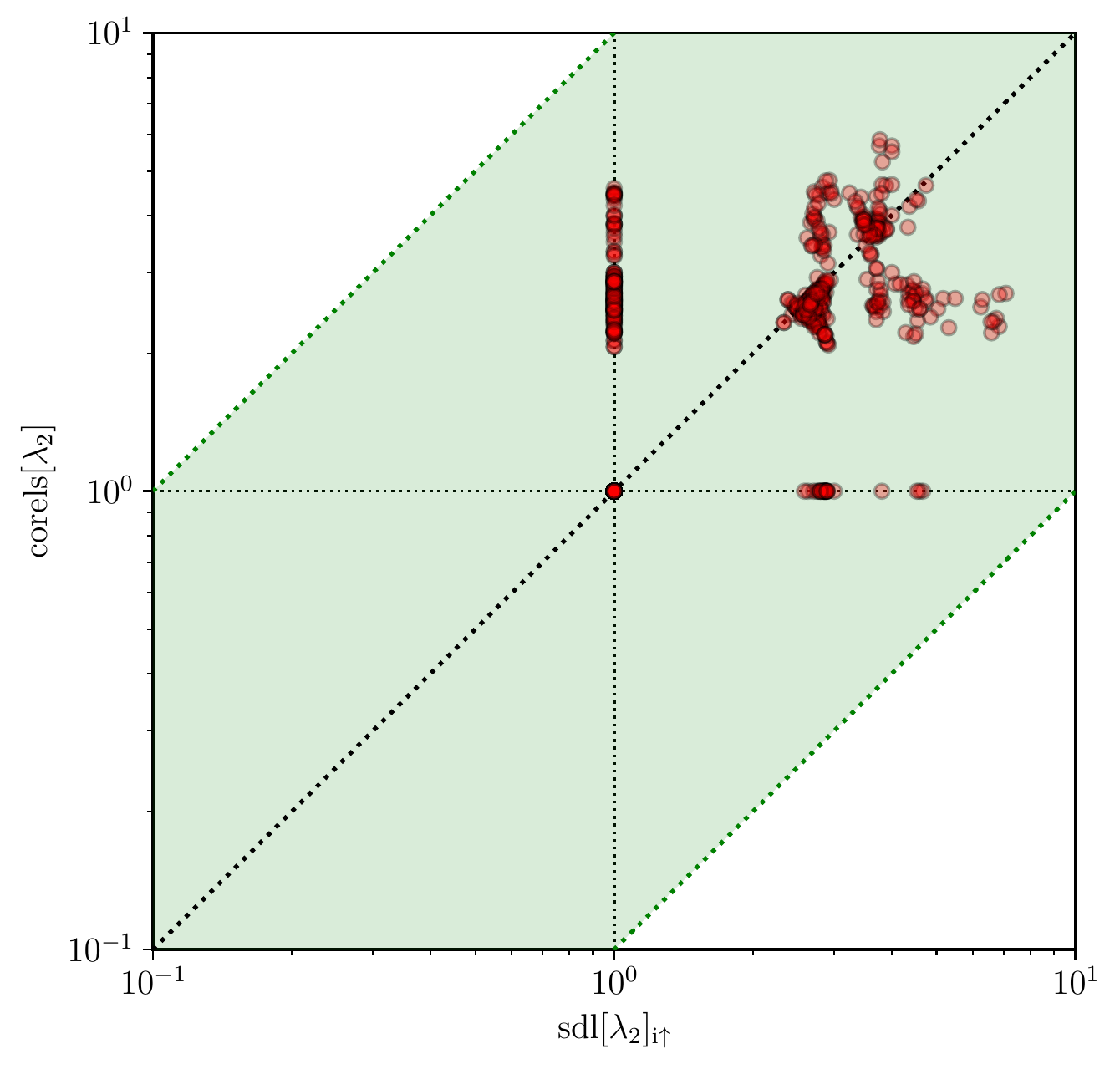}
    \caption{Explanations: sdl$[\lambda_2]_{i\uparrow}$ vs.~corels$[\lambda_2]$}
    \label{fig:sxsize-corels}
  \end{subfigure}%
  \begin{subfigure}[b]{0.3\textwidth}
    \centering
    \includegraphics[width=\textwidth]{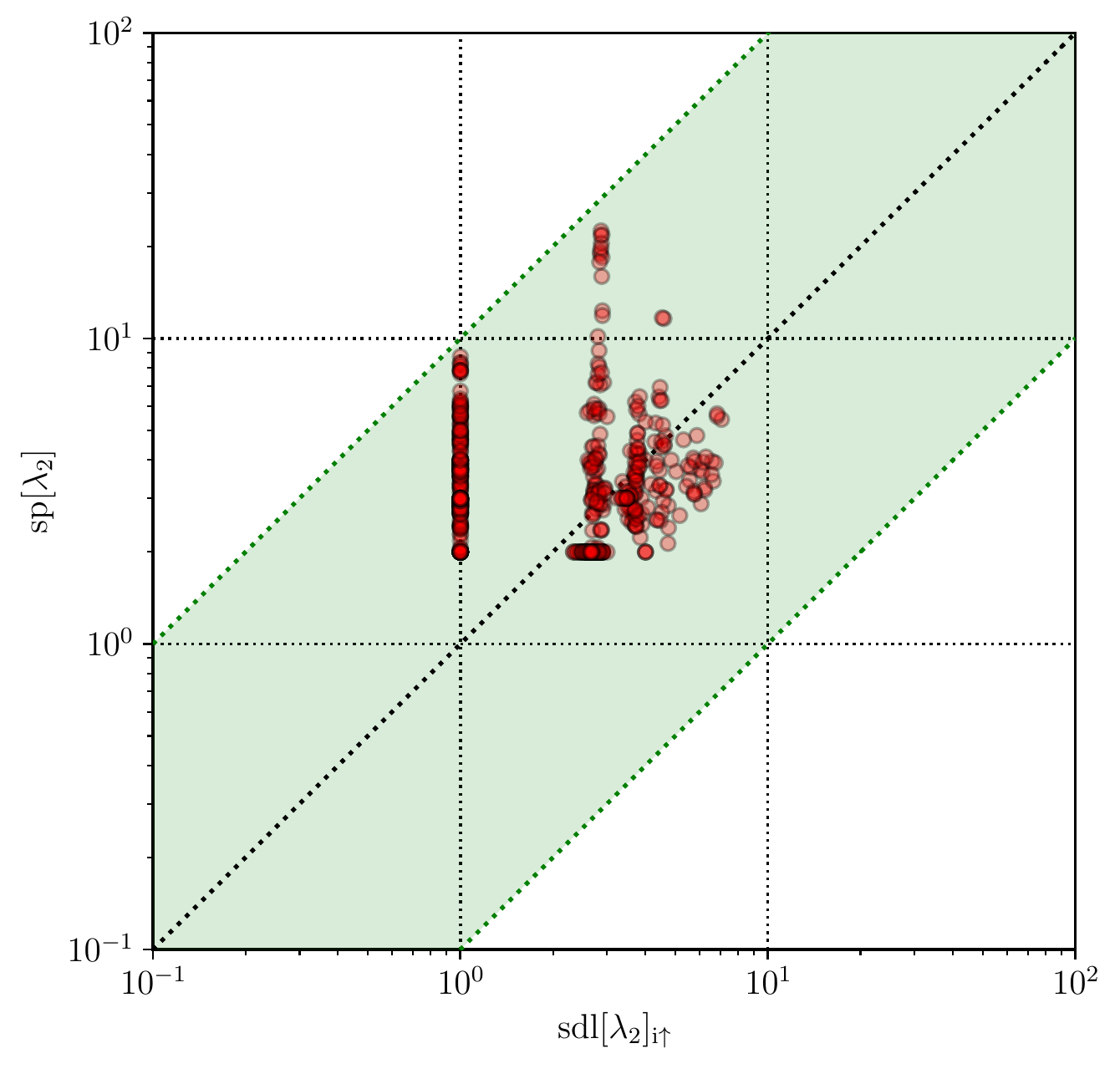}
    \caption{Explanations: sdl$[\lambda_2]_{i\uparrow}$ vs.~sp$[\lambda_2]$}
    \label{fig:sxsize-dset}
  \end{subfigure}
  \caption{Comparison of sparse models.}
  \label{fig:sparse}
\end{figure*}

This section describes the results of experimental assessment of the proposed approach to perfect and sparse decision lists and compares it with the state-of-the-art SAT-based decision sets~\cite{ipnms-ijcar18,yisb-corr20} as well as the only previous approach to optimal sparse decision lists we are aware of~\cite{rudin-kdd17a,rudin-jmlr17}.
Experimental results are obtained on the StarExec cluster\footnote{https://www.starexec.org/}~\cite{stump-ijcar14}, each computing node of which uses an Intel Xeon E5-2609~2.40GHz CPU with 128GByte of RAM, running CentOS~7.7.
The time limit and memory limit used per process are 1800 seconds and 16 GB.

For the evaluation, we use the benchmark suite previously studied in~\cite{yisb-corr20}.
Thus, the 71 datasets studied come from the UCI Machine Learning Repository~\cite{uci} and Penn Machine learning Benchmarks~\cite{pennml}.
We also use 5-fold cross validation, which results in 355 pairs of training and test data split with the ratio 80\% and 20\%, respectively.
Finally, feature domains are quantized into 2, 3, and 4 intervals and then \emph{one-hot encoded}~\cite{scikitlearn-full}.
The number of one-hot encoded features (training instances, resp.) per dataset in the benchmark suite varies from 3 to 384 (from 14 to 67557, resp.).
The total number of benchmark datasets is 1065 ($\text{71} \times \text{5} \times \text{3}$).

\subsubsection{Implementation} \label{sec:setup}

All the models proposed in \autoref{sec:sat}, and \autoref{sec:sep} are implemented as a set of Python scripts and solving is done by instrumenting calls to exact MaxSAT solver RC2-B~\cite{imms-sat18,imms-jsat19}.
The underlying SAT solver is Glucose~3~\cite{audemard-sat13}.
%
%
The complete MaxSAT model is referred to as $pdl_{\cup}$.
As was shown in \autoref{sec:sep}, separated models do not guarantee optimality of the size of decision list, and so we tested various ordering of the classes when computing separated decision lists.
Concretely, $pdl_{i\uparrow}$ and $pdl_{i\downarrow}$ refer to the separated models that order the classes by the increasing/decreasing number of training data in the classes.
Sparse models are referred to as $sdl[\lambda]_\circ$, where $\lambda$ is a regularized cost and ordering $\circ$ is from $\{\cup, i\downarrow, i\uparrow, a\downarrow, a\uparrow, c\downarrow, c\uparrow\}$ meaning that decision list computation is integrated, or done separately with the classes being ordered based on the increasing/decreasing number/accuracy/cost of training data in the classes, as defined in \autoref{sec:sep}.

\subsection{Perfect Models} \label{sec:rperfect}

We compare our prototype against state of the art in perfect decision set methods~\cite{ipnms-ijcar18,yisb-corr20}, namely $mds_2$, $mds_2^\star$, and $opt$.
While $mds_2$ generates a decision set with the smallest number of rules and $opt$ minimizes the number of literals, $mds_2^\star$ does rule minimization followed by literal minimization.
The comparison of perfect models is illustrated in \autoref{fig:perfect}.

\subsubsection{Performance.} \label{sec:ptime}

The performance of the perfect models is shown in \autoref{fig:prtime}.
As can be seen, $mds_2$ outperforms all the other rivals and trains 579 models.
This should not come as a surprise since $mds_2$ minimizes the number of rules.
It is followed by $mds_2^\star$, which sequentially applies rule and literal minimization -- $mds_2^\star$ can solve 399 benchmarks.
The best performing decision list model $pdl_{i\uparrow}$ comes third with 395 datasets handled successfully.
The optimal decision set approach $opt$ solves 350 instances.
Finally, $pdl_{i\downarrow}$ and $pdl_{\cup}$ can train 340 and 283 decision lists, respectively.

\subsubsection{Test Accuracy.} \label{sec:paccy}

Test accuracy computed for the benchmarks solved by all the competitors is shown in the cactus plot of \autoref{fig:paccy}.
Concretely, the plot depicts the value of \emph{test error} in percent.
On average, all the approaches perform similarly here and have test accuracy $\approx\text{80}\%$.
This is not surprising as all of them target perfectly accurate models.

\subsubsection{Model Size.} \label{sec:pmsize}

The model size calculated as the total number of literals in the model is shown in \autoref{fig:pmsize}.
Observe that optimal perfect decision lists $pdl_{\cup}$ are the smallest among all the approaches with the average size being 9.1 per model.
The second best model is $pdl_{i\uparrow}$ with 10.4 literals per model.
Note that the smallest size decision sets obtained by $opt$ have 23.3 literals on average.
The largest models are of $mds_2$ with 32.7 literals per model on average.
The pairwise comparison of model size for $opt$ and $pdl_{i\uparrow}$ is detailed in the scatter plot of \autoref{fig:pmsize-det}, which clearly demonstrates that perfect smallest size decision sets are usually larger than 
decision lists even when these are not guaranteed to be smallest in size.

\subsubsection{Average Explanation Size.} \label{sec:pxsize}

Although decision lists are smaller, the advantage of perfect decision sets is clearly the average explanation size per instance, which is calculated as described in \autoref{sec:size}.
This data is shown in \autoref{fig:pxsize}.
For instance, it takes 3.3 literals on average to explain a prediction of decision sets produced by $opt$.
For $mds_2^\star$ and $mds_2$ the numbers are $3.4$ and $5.1$, respectively.
Explanations for decision lists are larger; the best result is shown by $pdl_{\cup}$, which has 7.0 literals per explanation.
The best performing decision list model $pdl_{i\uparrow}$ has 9.3 literals per explanation.
The detailed comparison of the average explanation size for $opt$ and $pdl_{i\uparrow}$ is shown in the scatter plot of \autoref{fig:pxsize-det}.

\subsection{Sparse Models} \label{sec:rsparse}

The second part of our evaluation compares sparse models.
Here, the proposed approach is compared against sparse versions of decision sets $sp$ and $mds_2$ of~\cite{yisb-corr20} and 
optimal sparse decision lists produced by $corels$~\cite{rudin-kdd17a,rudin-jmlr17}.
Although we tested 3 values for regularized cost $\lambda\in\{\text{0.005}, \text{0.05}, \text{0.5}\}$, we report the results only for $\lambda_2=\text{0.05}$.
As~\cite{yisb-corr20} showed, the best trade-off for sparse decision sets was obtained for $\lambda_2=\text{0.05}$ and $\lambda_3=\text{0.5}$.
However, decision lists obtained for $\lambda_3$ are usually too sparse as they end up having a single rule predicting a constant class.
Therefore, hereinafter, the results are reported for configurations $sdl[\lambda_2]_\ast$ as well as for $sp[\lambda_2]$, $mds_2[\rho_1]$, $mds_2^\star[\rho_1]$, and $corels[\lambda_2]$.
(Note that the value of regularized cost $\rho_1=\text{0.05}$ is also taken from~\cite{yisb-corr20} unchanged. As $mds_2$ and $mds_2^\star$ minimize the number of rules, regularized cost $\rho_1$ is applied wrt.~the number of rules, which contrasts $\lambda_2$ applied to the number of literals.)
The results are shown in \autoref{fig:sparse}.

\subsubsection{Performance.} \label{sec:stime}

As can be observed in \autoref{fig:srtime}, $corels_{\lambda_2}$ is the fastest among the approaches for sparse models.
It solves 1016 benchmarks.
Sparse decision sets can be trained by $sp[\lambda_2]$ for 898 datasets while decision lists can be trained by $sdl[\lambda_2]_{i\uparrow}$ for 827 of them.
Observe that class ordering $i\uparrow$ based on the increasing number of instances per class outperforms the other configurations of $sdl[\lambda_2]$, which can tackle $\approx\text{800}$ datasets each.
The decision set competitors $mds_2[\lambda_2]$ and $mds_2^\star[\lambda_2]$ solve 772 benchmarks.
Finally, aggregated computation of smallest decision lists of $sdl[\lambda_2]_{\cup}$ handles 688 datasets.

\subsubsection{Test Accuracy.} \label{sec:saccy}

Although $corels[\lambda_2]$ outperforms its rivals in time, the accuracy of its decision lists is not the best.
The scatter plot in \autoref{fig:saccy-corels} depicts the value of test error $e=100\% - a$, where $a$ is test accuracy, for  $corels[\lambda_2]$ and $sdl[\lambda_2]_{i\uparrow}$.
Observe that in many cases the accuracy of $sdl[\lambda_2]_{i\uparrow}$ is significantly higher than of $corels[\lambda_2]$: the average accuracy of $corels[\lambda_2]$ is 40.2\% while the average accuracy of $sdl[\lambda_2]_{i\uparrow}$ is 69.9\%.
This clearly suggests that the sparsity measure used in our work enables us to train more accurate decision lists.
Also, as shown in \autoref{fig:saccy-dset}, the accuracy of $sdl[\lambda_2]_{i\uparrow}$ is on par with the accuracy of sparse decision sets of $sp[\lambda_2]$, which on average equals 67.6\%.

\subsubsection{Model Size.} \label{sec:smsize}

As detailed in \autoref{fig:smsize}, the smallest models are obtained with sparse decision lists of $corels[\lambda_2]$ and $sdl[\lambda_2]_{\ast}$.
The average number of literals in the lists produced by $corels[\lambda_2]$, $sdl[\lambda_2]_{\cup}$, and $sdl[\lambda_2]_{i\uparrow}$ is 2.7, 2.3, and 2.4, respectively (these numbers are calculated across the instances \emph{solved} by the corresponding tools).
Similar results are demonstrated by the other configurations of $sdl[\lambda_2]_\ast$.
In contrast, the average size of sparse decision sets of $sp[\lambda_2]$, $mds_2[\lambda_2]$, and $mds_2^\star[\lambda_2]$ is 6.9, 22.5, and 17.9, respectively.

\subsubsection{Average Explanation Size.} \label{sec:sxsize}

\autoref{fig:sxsize-corels} and \autoref{fig:sxsize-dset} provide a comparison of $sdl[\lambda_2]_{i\uparrow}$ against $corels[\lambda_2]$ and $sp[\lambda_2]$ in terms of the average explanation size.
In contrast to the case of perfect models, an average explanation for decision lists of $sdl[\lambda_2]_{i\uparrow}$ has 2.1 literals while explanations of sparse decision sets of $sp[\lambda_2]$ are of size 3.4.
This suggests that sparse decision lists not only are smaller than sparse decision sets but they also provide a user with explanations that are more succinct.
The average explanation size of the decision lists of $corels[\lambda_2]$ is 2.3.
(The average numbers shown here are collected across all benchmarks solved by the corresponding tools.)

\section{Conclusion} \label{sec:conc}%

In this paper we develop SAT based methods to construct optimal perfect decision lists.  This is the first method we are aware of for optimal perfect decision lists. The method is extended to construct optimal (or near-optimal) sparse decision lists where we trade off accuracy for size. While existing bespoke methods for optimal sparse decision lists are considerably more scalable, interestingly the accuracy of the models they construct are lower, probably because the size measure we use is more fine grained. We provide the first comparison of decision sets and lists in terms of model size and explanation size.  For perfect models decision sets are preferable, but surprisingly this reverses for sparse models.

\bibliography{refs}

\end{document}